\documentclass[12pt]{iopart}

\usepackage{ifpdf}

\ifpdf
\usepackage[pdftex]{graphicx}
\usepackage{epstopdf}
\graphicspath{{./image/}}
\else
\fi

\usepackage{array}
\usepackage{rotating}

\usepackage{multirow}

\usepackage{multicol}

\usepackage{soul}

\usepackage[normalem]{ulem}
\usepackage{iopams}
\expandafter\let\csname equation*\endcsname\relax
\expandafter\let\csname endequation*\endcsname\relax
\usepackage[cmex10]{amsmath}
\usepackage{amssymb}

\usepackage{makecell}
\usepackage{textcomp}

\usepackage{lipsum}

\usepackage{capt-of}

\usepackage{tabularx}
\usepackage{bm}
\usepackage{calc}

\usepackage{xcolor,colortbl}

\usepackage{sidecap, caption}
\usepackage{tabu}
\usepackage{mdframed}

\usepackage{tikz}
\usetikzlibrary{shapes.geometric}
\usetikzlibrary{calc}
\newcommand{\toppage}{
	\begin{tikzpicture}[remember picture,overlay]
	\node[align=left, anchor=north west]
	at ($(current page.north west) + (2,-1)$)
	{\texttt{This article has been published in \textit{Physiological Measurement}}};
	\end{tikzpicture}
}

\begin{document}
	
	\title[Personalized Automatic Sleep Staging with Single-Night Data]{Personalized Automatic Sleep Staging with Single-Night Data: a Pilot Study with KL-Divergence Regularization}
	
	\author{Huy Phan$^1$, Kaare Mikkelsen$^2$, Oliver Y. Ch\'{e}n$^3$, Philipp Koch$^4$, Alfred Mertins$^4$, Preben Kidmose$^2$, Maarten De Vos$^{3,5}$}
	
	\address{$^1$School of Electronic Engineering and Computer Science, Queen Mary University of London, UK \\
		$^2$Department of Engineering, Aarhus University, Denmark \\
		$^3$Institute of Biomedical Engineering, University of Oxford, UK \\
		$^4$Institute for Signal Processing, University of L\"ubeck, Germany \\
		$^5$Department of Electrical Engineering, KU Leuven, Belgium}
	\ead{h.phan@qmul.ac.uk}
	\vspace{10pt}
	\begin{indented}
		\item[]April 2020
	\end{indented}
	
	\begin{abstract}
		Brain waves vary between people. An obvious way to improve automatic sleep staging for longitudinal sleep monitoring is personalization of algorithms based on individual characteristics extracted from the first night of data. As a single night is a very small amount of data to train a sleep staging model, we propose a Kullback-Leibler (KL) divergence regularized transfer learning approach to address this problem. We employ the pretrained SeqSleepNet (i.e. the subject independent model) as a starting point and finetune it with the single-night personalization data to derive the personalized model. This is done by adding the KL divergence between the output of the subject independent model and the output of the personalized model to the loss function during finetuning. In effect, KL-divergence regularization prevents the personalized model from overfitting to the single-night data and straying too far away from the subject independent model. Experimental results on the Sleep-EDF Expanded database with 75 subjects show that sleep staging personalization with a single-night data is possible with help of the proposed KL-divergence regularization. On average, we achieve a personalized sleep staging accuracy of $79.6\%$, a Cohen's kappa of $0.706$, a macro F1-score of $73.0\%$, a sensitivity of $71.8\%$, and a specificity of $94.2\%$.
		We find both that the approach is robust against overfitting and that it improves the accuracy by $4.5$ percentage points compared to non-personalization and $2.2$ percentage points compared to personalization without regularization.
		\toppage
	\end{abstract}
	
	%
	%
	%
	%
	%

	\section{Introduction}
	
	The increased awareness of the important role of sleep in protecting our mental and physical health \cite{Siegel2005} has been translated in an increased demand in personal sleep monitoring tools. For such purpose, automating sleep scoring is vital and indispensable since manual scoring is simply too expensive, time-consuming, and labor-intensive \cite{Iber2007, Hobson1969}. The advance of machine learning, deep learning in particular, coupled with the availability of large sleep databases \cite{nsrr2019,Oreilly2014,Stephansen2018} has stimulated a new wave of interest in developing automatic sleep staging methods. In fact, machine performance in sleep staging has progressed significantly, being on par with manual scoring by sleep experts, thanks to recent methods based on deep learning \cite{Phan2019e, Supratak2017, Stephansen2018}. 
	
	\begin{figure} [!t]
			\centering
			\includegraphics[width=0.85\linewidth]{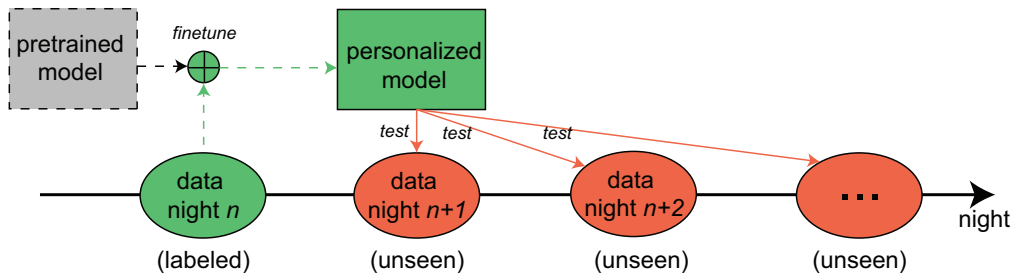}
			\caption{Personalization with single-night data: a pretrained model is finetuned with the labeled data of night $n$ of an undividual and yields the personalized model which is tested on the same individual's unseen data of nights $n+1$, $n+2$, $\ldots$} 
			\label{fig:personalization}
	\end{figure}
	
	The above-mentioned state-of-the-art classification performance is only possible using supervised learning.
	That is, we need data to be recorded and manually labeled from a cohort of subjects, followed by model training based on the labeled data. In fact, the recent expert-level performance is only obtainable with a large cohort (i.e. hundreds or thousands of subjects) \cite{Phan2019e, Stephansen2018}. Collecting and manually scoring a large amount of sleep data is a vast burden, particularly for wearable EEG devices like in-ear EEG \cite{Mikkelsen2019b} or around-the-ear EEG \cite{Mikkelsen2019, Sterr2018}, 
	in which case the work load is increased by the need for an added PSG for reference.
	Utilizing and including available sleep data for training a sleep staging algorithm in novel settings is not easy, due to channel mismatch caused by differences in channel layouts, electrode placements, recording devices and software, preprocessing procedure, normalization parameters, clinical cohort characteristics, etc. \cite{Phan2019e}. The work in \cite{Phan2019f,Phan2019e} proposed a transfer learning approach to circumvent the above-mentioned channel mismatch and enable knowledge transfer from a large dataset to a small cohort, making a deep learning model for a different, specific setting with low amount of data possible. However, such a transfer learning approach still requires data from a dozen of subjects to succeed. Although collecting and labeling this relatively small amount of sleep data would not be a big problem, here we want to push this data constraint to its extreme and question whether it is possible to adapt a pretrained model with single-night data of a particular subject, i.e. personalization, even without knowing in which setting the data is recorded. By personalization, we mean the parameters of the pretrained model are adapted to an individual's data to convert into a personalized model which is later tested on the same individual's future unseen data as illustrated in Figure \ref{fig:personalization}. If personalization with single-night data in an unknwon setting is possible, it would be convenient for one to build a model for personalized sleep monitoring using his/her very own minimal data recorded with a particular device. It is equally important and necessary when privacy and security become a serious concern \cite{Agarwal2019, Martinovic2012, Bonaci2015}, and thus, owning EEG data from others to form a cohort for transfer learning \cite{Phan2019e} would be more and more difficult. An additional and very important benefit of personalization is that it has previously been shown that automatic sleep scoring becomes more accurate when the classifier can focus on the peculiarities of the individual (see \cite{Mikkelsen2017} and [9]). This is especially the case when using non-standard EEG montages, for instance in in-ear EEG and around-the-ear EEG. It should be noted that this personalization problem is different from that in \cite{Mikkelsen2018} in which a cohort of subjects is known and a model is trained on the cohort before personalizing for a subject in the same cohort. Here, we assume there is no information about the cohort or recording settings but only a single-night data of a target subject is available.

	Building a deep-learning model using  single-night data is challenging. First, the model can easily overfit the data regardless of whether we train a model from scratch or finetune a pretrained model \cite{Phan2019e}. Second, different subjects are expected to have varying convergence/overfitting rate when training/finetuning the personalized model. Therefore, we do not know when the model will start overfitting, as we do not have validation data at hand for model selection as in the case of a cohort \cite{Phan2019e,Phan2019f}. Third, regular data normalization cannot be done as a cohort's statistics are unknown. In this work, we take on this `personalization with single-night data' challenge and propose an approach based on transfer learning to deal with it. We employ the pretrained SeqSleepNet \cite{Phan2019e,Phan2019a} (i.e. the subject independent (SI) model), and finetune it with single-night data from a single subject from an unknown cohort to accomplish personalization. Note that, the source-domain cohort which was used for pretraining the model is also assumedly unknown. To remedy the overfitting problem, KL-divergence between the output of the SI model and the personalized model is introduced to regularize the network. The KL-divergence regularization, in effect, prevents the personalized model from drifting too far away from the SI model. 
	Once we get rid of overfitting, model selection is no longer an issue as we can keep finetuning the SI model as long as we need. Experiments on 75 subjects of the Sleep-EDF Expanded database \cite{Kemp2000, Goldberger2000} show that KL-divergence regularized personalization with single-night data is robust against overfitting and achieves an average sleep staging accuracy of $79.6\%$, improving $4.5$ and $2.2$ percentage points over non-personalization and personalization without KL-divergence regularization, respectively.
	
	\section{Material}
	
	We used the Sleep-EDF Expanded database (Sleep Cassette subset, version 2018)  \cite{Kemp2000, Goldberger2000} in this study. This database consists of 78 healthy Caucasian subjects aged 25-101. It is particularly suitable for this study as there are 75 out of 78 subjects with two subsequent day-night PSG recordings collected for each. Three subjects (subjects 13, 36, and 52) whose one recording was lost due to device failure were excluded from the personalization experiments. Manual scoring was done by sleep experts according to R\&K standard \cite{Hobson1969} and each 30-second PSG epoch was labeled as one of eight categories {W, N1, N2, N3, N4, REM, MOVEMENT, UNKNOWN}. We merged N3 and N4 into a single stage N3 and excluded MOVEMENT and UNKNOWN categories as in previous experiments in earlier versions of the database \cite{Imtiaz2015, Tsinalis2016, Tsinalis2016b, Supratak2017, Phan2019b}. We used the Fpz-Cz EEG channel sampled at 100 Hz in this study. As different portions of this database have been used in the literature, it should be stressed that we only made use of the \emph{in-bed} parts (from \emph{lights off} time to \emph{lights on} time) recommended in \cite{Imtiaz2014,Imtiaz2015} and adopted in many existing works \cite{Tsinalis2016, Tsinalis2016b, Phan2019b, Phan2019e, phan2018c, phan2018d, Mikkelsen2018, Andreotti2018}. 
	
	\section{Methods}
	
	\subsection{Sequence-to-Sequence Sleep Staging with SeqSleepNet}
	
	\emph{SeqSleepNet}, recently proposed in \cite{Phan2019a},  has demonstrated state-of-the-art performance on several sleep databases \cite{Phan2019a,Phan2019f} and its suitability for transfer learning tasks \cite{Phan2019f,Phan2019e}. We employ it in this work to study sleep-staging personalization. As a sequence-to-sequence sleep-staging model \cite{Phan2019a}, SeqSleepNet learns to maximize the conditional probability $p(\mathbf{y}_1, \mathbf{y}_2, \ldots, \mathbf{y}_L \,|\, \mathbf{S}_1, \mathbf{S}_2, \ldots, \mathbf{S}_L)$ \cite{Phan2019a}. In other words, it receives a sequence of $L$ consecutive epochs $(\mathbf{S}_1, \mathbf{S}_2, \ldots, \mathbf{S}_L)$ and classifies them at once into a sequence of corresponding sleep stages $(\mathbf{y}_1, \mathbf{y}_2, \ldots, \mathbf{y}_L)$, where $\mathbf{y}$ is a one-hot encoding vector.

	\begin{figure} [!t]
		\centering
		\includegraphics[width=0.75\linewidth]{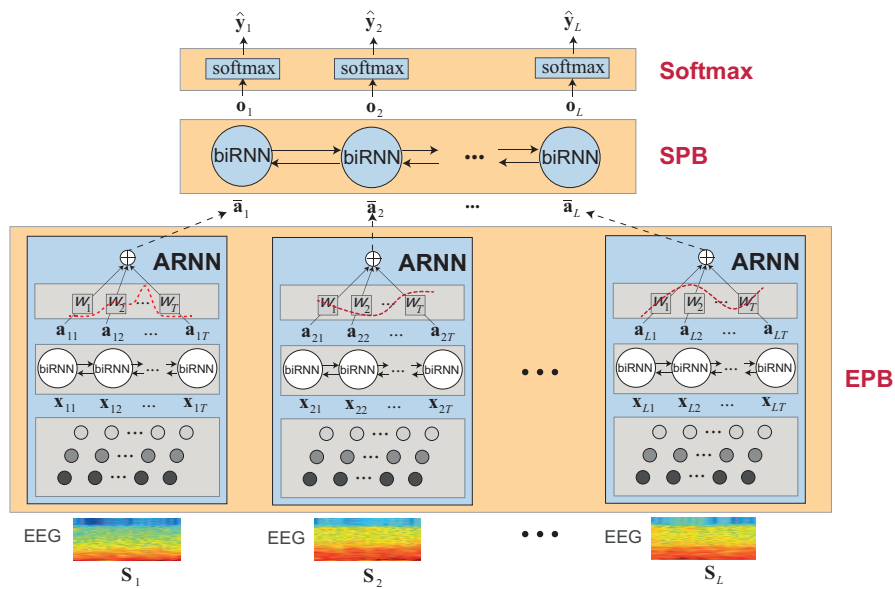}
		\caption{Illustration of SeqSleepNet which is composed of three components: epoch processing block (EPB), sequence processing block (SPB), and Softmax. Image adapted from \cite{Phan2019a}.} 
		\label{fig:seqsleepnet}
	\end{figure}
	
	To be fed into the network, the EEG signal of a 30-second epoch is transformed into a time-frequency image $\mathbf{S} \in \mathbb{R}^{F\times T}$ obtained via short-time Fourier transform (STFT), where $F$ is the number of frequency bins and $T$ is the number of time instances (cf. Section \ref{ssec:experimental_setup}). The network is composed of three main components: epoch processing block (EPB), sequence processing block (SPB), and Softmax, as illustrated in Figure \ref{fig:seqsleepnet}.
	
	{\bf EPB.} EPB is essentially an attention-based RNN (ARNN) \cite{phan2018d} that is shared by all epochs in the input sequence for short-term (i.e. intra-epoch) sequential modelling. The ARNN subnetwork consists of a \emph{filterbank} layer \cite{phan2018c}, a bidirectional RNN realized by long short-term memory (LSTM) cells \cite{Hochreiter1997} with recurrent batch normalization \cite{Cooijmans2016}, and a self-attention layer \cite{Luong2015b}. The trainable filterbank layer with $M$ filters is designed to smooth and reduce the frequency dimension of each epoch $\mathbf{S}$ from $F$ to $M$, where $M < F$ \cite{phan2018c}. The resulting image is then treated as a sequence of $T$ local feature vectors $(\mathbf{x}_1, \mathbf{x}_2, \ldots, \mathbf{x}_T)$ (corresponding to $T$ spectral columns) which is encoded by the bidirectional RNN into a sequence of output vectors $(\mathbf{a}_1, \mathbf{a}_2, \ldots, \mathbf{a}_T)$. The self-attention layer \cite{Luong2015b} is trained to produce attention weights $(w_1, w_2, \ldots, w_T)$ and combines the output vectors into a single feature vector $\bar{\mathbf{a}} = \sum\nolimits^{T}_{t=1}w_t\mathbf{a}_t$ to represent the epoch $\mathbf{S}$.
	
	{\bf SPB.} SPB is a bidirectional RNN for long-term (i.e. inter-epoch) sequential modelling. Similar to the RNN in EPB, this RNN is also realized by LSTM cells  \cite{Hochreiter1997} with recurrent batch normalization \cite{Cooijmans2016}. After EPB, the input sequence $(\mathbf{S}_1, \mathbf{S}_2, \ldots, \mathbf{S}_L)$ has been converted into a sequence of feature vectors $(\bar{\mathbf{a}}_1, \bar{\mathbf{a}}_2, \ldots, \bar{\mathbf{a}}_T)$. In turn, the bidirectional RNN iterates over the sequence of induced feature vectors and encode it into the sequence of output vectors $(\mathbf{o}_1, \mathbf{o}_2, \ldots, \mathbf{o}_L)$.
	
	{\bf Softmax.} Given the sequence of output vectors $(\mathbf{o}_1, \mathbf{o}_2, \ldots, \mathbf{o}_L)$, classification eventually takes place at the Softmax component to produce the sequence posterior probabilities  $(\hat{\mathbf{y}}_1, \hat{\mathbf{y}}_2, \ldots, \hat{\mathbf{y}}_L)$, where $\hat{\mathbf{y}}_l$ coresponds to the epoch at index $l$, $1 \le l \le L$, in the input sequence. Similar to the SeqSleepNet+ variant in \cite{Phan2019e}, the softmax layer is shared between all epochs.  
	
	The network is trained end-to-end to minimize the sequence classification loss over all $N$ training sequences in the training data:
	\begin{align}
	E(\Theta) &= -\frac{1}{L}\sum_{n=1}^{N}\sum_{l=1}^{L} \mathbf{y}_{nl}\log\left(\mathbf{\hat{y}}_{nl}\left(\Theta\right)\right) + \frac{\lambda}{2}\|\Theta\|^2_2 \nonumber\\
	&= -\frac{1}{L}\sum_{n=1}^{N}\sum_{l=1}^{L}\sum_{c \, \in \, \mathcal{C}} \mathbb{I}(y_{nl}=c)\log P_\Theta(\hat{y}_{nl} = c) + \frac{\lambda}{2}\|\Theta\|^2_2,
	\label{eq:sequence_loss}
	\end{align}
	where $\mathcal{C} = \{\text{W}, \text{N1}, \text{N2}, \text{N3}, \text{REM}\}$ is the set of all possible sleep stages. In (\ref{eq:sequence_loss}), $\mathbb{I}(\cdot)$ is the indicator function, $y_{nl}$ and $\hat{y}_{nl}$ denotes the ground-truth and output discrete labels of the $l^{\text{th}}$ epoch in the $n^{\text{th}}$ sequence, respectively. $\Theta$ denotes the trainable parameters of the network and $\lambda$ is the coefficient of the $\ell_2$-norm regularization term.
	
	
	\subsection{KL-Divergence Regularization for Personalization}
	
\begin{SCfigure}[][t]
		\includegraphics[width=0.45\linewidth]{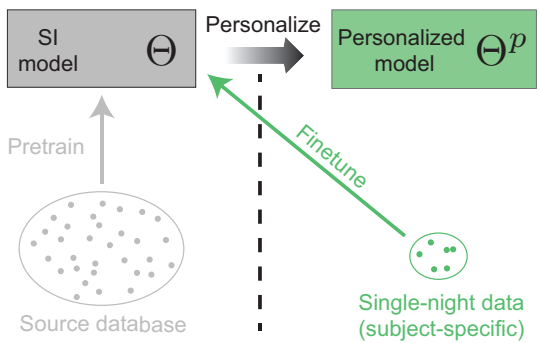}
		\captionsetup{width=1\textwidth}
		\caption[]{Illustration of sleep personalization with single-night data. The subject independent (SI) model $\Theta$, which is pretrained with a source-domain database (assumedly unknown), is finetuned on the single-night data of a target subject to derive the personalized model $\Theta^p$.}
		\label{fig:sleep_personalization}
	\end{SCfigure}

	Given the small amount of data (one night) it is not feasible to train a deep learning model like SeqSleepNet from scratch. 
	As mentioned before, we, therefore, pursue a transfer learning approach similar to \cite{Phan2019e,Phan2019f} for personalization. We use the pretrained SeqSleepNet model from \cite{Phan2019e}, which was pretrained using the C4-A1 EEG data from 200 subjects (686,610 epochs in total) of the Montreal Archive of Sleep Studies (MASS) database \cite{Oreilly2014} (i.e. the source database), as the subject independent (SI) model denoted by $\Theta$. We would like to remind the reader that the MASS cohort is assumedly unknown here. The SI model $\Theta$ then serves as the starting point and is finetuned using the single-night data of a target subject to derive the personalized model, denoted by $\Theta^p$, as illustrated in Figure \ref{fig:sleep_personalization}. Note that channel mismatch is expected between the source-domain MASS database and the target subject's personalization data, and finetuning is supposed to address both channel mismatch and personalization. We investigate four finetuning strategies \{\emph{All}, \emph{EPB+Softmax}, \emph{SPB+Softmax}, \emph{Softmax}\} similar to those in \cite{Phan2019e,Phan2019f}. When components of the pretrained network (i.e. the entire network, EPB+Softmax, SPB+Softmax, or Softmax depending on the finetuning strategies) are finetuned, their weights are adapted with the personalization data while the rest remains fixed.
	
	
	The study in \cite{Phan2019e} showed that sleep transfer learning requires roughly at least ten subjects' data, leaving personalization with the single-night data of a target subject exposed to the substantial risk of overfitting. In fact, we experimentally see that the personalized model tends to overfit the personalization data very easily. Moreover, there exists no viable way to select the right model during finetuning before overfitting starts. One may leave out a portion of the one-night data for validation. However, since the validation data is distributed very similarly to the finetuning data, this leave-out validation data is also overfitted easily and cannot be used to identify overfitting. To remedy overfitting, we propose to regularize the sequential classification loss function in (\ref{eq:sequence_loss}) with the KL divergence between the posterior probability outputs of the SI model $\Theta$ and the ones from the personalized model $\Theta^p$, which constrains the personalized model not to stray too far away from the SI model \cite{Yu2013}. Given an input sequence $(\mathbf{S}_1, \mathbf{S}_2, \ldots, \mathbf{S}_L)$, KL divergence between the outputs of the two models reads:
	\begin{align}
	D_{\text{KL}} = \frac{1}{L}\sum_{l=1}^L\sum_{c \, \in \, \mathcal{C}} P_{\Theta}(\hat{y}_l = c)\log\left(\frac{P_{\Theta}(\hat{y}_l = c)}{P_{\Theta^p}(\hat{y}_l =c)}\right).
	\end{align}
	
	\sidecaptionvpos{figure}{c}
	
	The KL-divergence regularization is added into the sequential classification loss function in (\ref{eq:sequence_loss}) to form the loss function for personalization:
	\begin{align}
	E(\Theta^p) = &-(1-\alpha)\frac{1}{L}\sum_{n=1}^{N}\sum_{l=1}^{L}\sum_{c\,\in\,\mathcal{C}}\mathbb{I}(y_{nl} = c)\log P_{\Theta^p}(\hat{y}_{nl} = c) + \frac{\lambda}{2}\|\Theta^p\|^2_2 \nonumber \\ &+  \alpha\frac{1}{L}\sum_{n=1}^{N}\sum_{l=1}^L\sum_{c \, \in \, \mathcal{C}} P_{\Theta}(\hat{y}_{nl} = c)\log\left(\frac{P_{\Theta}(\hat{y}_{nl} = c)}{P_{\Theta^p}(\hat{y}_{nl} =c)}\right),
	\label{eq:regularized_sequence_loss}
	\end{align}
	where $\alpha \in [0,1]$ is the KL-divergence regularization coefficient, regulating how far the personalized model $\Theta^p$ deviates from the SI model $\Theta$. When $\alpha = 0$, the KL-divergence regularization is cancelled out and the personalization turns out to be the same as regular finetuning in \cite{Phan2019e,Phan2019f}. In this case, the pretrained SI model is adapted solely on the personalization data. In contrast, when $\alpha=1$, we trust the pretrained SI model completely and ignore all the new information of the personalization data. Since the term $\alpha\frac{1}{L}\sum\limits_{n=1}^N\sum\limits_{l=1}^L\sum\limits_{c\,\in\,\mathcal{C}} P_{\Theta}(\hat{y}_{nl} =c)\log P_{\Theta}(\hat{y}_{nl} = c)$ in the KL-divergence regularization term in (\ref{eq:regularized_sequence_loss}) does not depend on the personalized network $\Theta^p$, the KL-divergence regularized loss function can be simplified as:
	\begin{align}
	E'(\Theta^p) = &-(1-\alpha)\frac{1}{L}\sum_{n=1}^{N}\sum_{l=1}^{L}\sum_{c\,\in\,\mathcal{C}} \mathbb{I}(y_{nl} = c)\log P_{\Theta^p}(\hat{y}_{nl} = c) + \frac{\lambda}{2}\|\Theta^p\|^2_2 \nonumber \\ &-  \alpha\frac{1}{L}\sum_{n=1}^{N}\sum_{l=1}^L\sum_{c\,\in\,\mathcal{C}} P_{\Theta}(\hat{y}_{nl} = c)\log P_{\Theta^p}(\hat{y}_{nl} = c).
	\label{eq:simplified_regularized_sequence_loss}
	\end{align}
	
	It turns out that the loss function for personalization in (\ref{eq:simplified_regularized_sequence_loss}) consists of two terms: (1) the cross-entropy between the output of the personalized model $\Theta^p$ and the ground-truth, and (2) the cross-entropy between the output of the personalized model $\Theta^p$ and the output of the pretrained SI model $\Theta$. As a result, model personalization  is equivalent to changing the target distribution from the unknown source-domain database (the MASS database used for pretraining) to a linear interpolation of the source-domain data distribution and the personalized data distribution \cite{Yu2013}. This interpolation prevents the network from overfitting the personalization data. 
	
	\section{Experimental Setup}
	\label{ssec:experimental_setup}
	
	For each of the 75 subjects with two day-night recordings of the Sleep-EDF Expanded database, we conducted finetuning of the pretrained SeqSleepNet \cite{Phan2019e} using the data from the first night and evaluating the personalized model on the data from the second night. We experimented with different values for the KL-divergence regularization coefficient $\alpha$ in the set \{0, 0.2, 0.4, 0.6, 0.8\} to investigate its influence. Note that, when $\alpha = 0$, we excluded the KL-divergence regularization completely. This case is considered the baseline for comparison with the proposed approach.
	
	The EEG signal was divided into 30-second epochs. Each epoch was transformed into a log-magnitude time-frequency image by the following procedure: the signal was divided into two seconds windows with 50\% overlap, multiplied with a Hamming window, transformed to the frequency domain by means of a 256-point Fast Fourier Transform (FFT), and the amplitude spectrum was log-transformed. This resulted in an image of size $F \times T$ where $F = 129$ (the number of frequency bins) and $T = 29$ (the number of spectral columns).
	
	
	\section{Results}
	\subsection{SeqSleepNet's performance on regular training setting.}
	\label{ss:regular_experiment}
	\setlength\tabcolsep{2.25pt}
	\begin{table}[!t]
		\caption{Performance on regular (scratch) training setup via 10-fold cross validation.}
		\footnotesize
		\begin{center}
			\begin{tabular}{|>{\arraybackslash}m{1.1in}|>{\centering\arraybackslash}m{1in}|>{\centering\arraybackslash}m{0.4in}|>{\centering\arraybackslash}m{0.4in}|>{\centering\arraybackslash}m{0.4in}|>{\centering\arraybackslash}m{0.4in}|>{\centering\arraybackslash}m{0.4in}|>{\centering\arraybackslash}m{0in} @{}m{0pt}@{}}
				\cline{1-7}
				\multirow{2}{*}{System} & \multirow{2}{*}{\makecell{Data portion}} & \multicolumn{5}{c|}{Overall metrics} & \parbox{0pt}{\rule{0pt}{0.25ex+\baselineskip}} \\ [0ex]  	
				
				\cline{3-7}
				& & Acc. & $\kappa$ & MF1 & Sens. &  Spec. &\parbox{0pt}{\rule{0pt}{0.25ex+\baselineskip}} \\ [0ex]  	
				\cline{1-7}
				\bf SeqSleepNet & \emph{in-bed} only & $79.1$ & $0.708$ & $74.6$  & $74.2$ & $94.2$&  \parbox{0pt}{\rule{0pt}{0.25ex+\baselineskip}} \\ [0ex]  	
				DeepSleepNet \cite{Supratak2017} & \emph{in-bed} only & $78.5$ & $0.702$ & $75.3$  & $75.0$ & $94.1$ &  \parbox{0pt}{\rule{0pt}{0.25ex+\baselineskip}} \\ [0ex]  	
				\hline
				\bf SeqSleepNet & \emph{in-bed} $\pm$ 30 min & $82.6$ & $0.760$ & $76.4$  & $76.3$ & $95.4$ &  \parbox{0pt}{\rule{0pt}{0.25ex+\baselineskip}} \\ [0ex]  	
				SleepEEGNet \cite{MousaviI2019} & \emph{in-bed} $\pm$ 30 min & $80.0$ & $0.730$ & $73.6$ & $-$ &  $-$ &  \parbox{0pt}{\rule{0pt}{0.25ex+\baselineskip}} \\ [0ex]  	
				\cline{1-7}
			\end{tabular}
		\end{center}
		\label{tab:scratch_performance}
	\end{table}
	\begin{figure} [!t]
			\centering
			\includegraphics[width=0.9\linewidth]{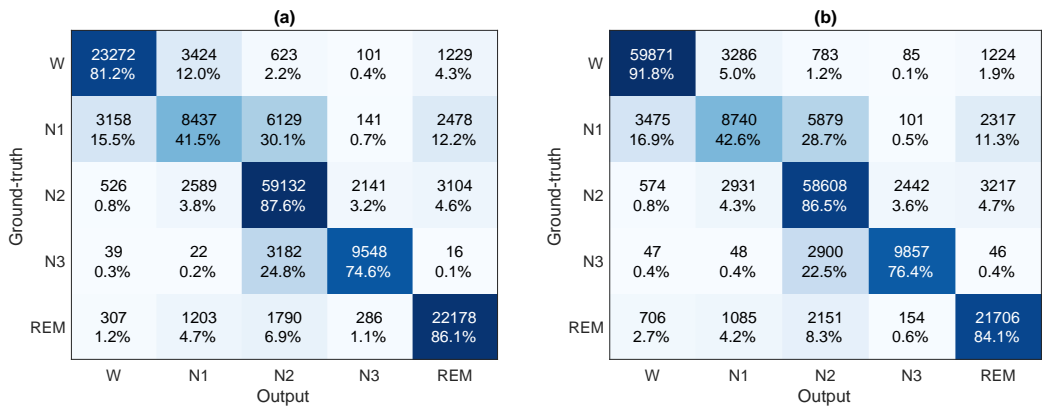}
			\caption{Confusion matrices obtained by SeqSleepNet. (a) \emph{in-bed} data only, (b) \emph{in-bed} data $\pm$ 30 min.} 
			\label{fig:scratch_confusion_matrices}
	\end{figure}
	
	SeqSleepNet \cite{Phan2019a,Phan2019e} requires the data to be normalized to zero mean and unit standard deviation \cite{Phan2019a,Phan2019e}. Unfortunately, in our case neither the source-domain cohort (i.e. the MASS cohort) nor the target subject's cohort (i.e. the Sleep-EDF cohort) are known. We, therefore, cannot normalize the personalization data using the cohort's statistics. In addition, we experimentally found that model personalization is sensitive to differences in magnitude of data between two nights, and per-subject data normalization resulted in poor performance in some subjects with such substaintial magnitude difference. To rule out this difference, we alternatively performed \emph{per-night normalization} in which data of one night recording was normalized by its mean and standard deviation.
	
	The implementation was based on the Tensorflow framework \cite{Abadi2016}. The pretrained SeqSleepNet was parametrized similarly to the one in \cite{Phan2019e} and used a sequence length of $L=20$.  For personalization, the pretrained SeqSleepNet was finetuned on the single-night finetuning data for 50 finetuning epochs and the performance was recorded every 5 finetuning epochs. Finetuning was performed using the Adam optimizer \cite{Kingma2015} with a learning rate of $10^{-4}$. 
	

	SeqSleepNet has been reported to achieve state-of-the-art performance on the MASS database \cite{Oreilly2014} (i.e. the source domain used for pretraining) and the earlier version of the Sleep-EDF Expanded database with 20 subjects \cite{Kemp2000, Goldberger2000}. It is worth assessing its performance on the experimental database on a regular (scratch) training setup. To this end, we conducted 10-fold cross validation on all 78 subjects. At each iteration, 7 subjects were left out for validation (i.e. model selection). During training, the network achieving the best overall accuracy on the validation subjects was retained for evaluation on the test subjects. The results of 10 cross-validation folds were pooled to calculate the overall metrics, including accuracy, macro F1-score (MF1) \cite{Yang1999}, Cohen’s kappa ($\kappa$) \cite{McHugh2012}, sensitivity, and specificity. Beside SeqSleepNet we also implemented the end-to-end variant of the popular DeepSleepNet \cite{Supratak2017,Phan2019a} for comparison. In addition, we include results for another common usage of the database in which 30 minutes of data before and after in-bed parts are additionally included. The performance is shown in Table \ref{tab:scratch_performance} in which SeqSleepNet not only obtains better performance than the DeepSleepNet counterpart but also outperforms the most recent results in \cite{MousaviI2019} on this latest version of the Sleep-EDF Expanded database. The accuracy of the sleep stages is also shown in the confusion matrices in Figure \ref{fig:scratch_confusion_matrices}.
	
	\subsection{Influence of KL-divergence regularization.}
	
\begin{figure} [!t]
		\centering
		\includegraphics[width=0.8\linewidth]{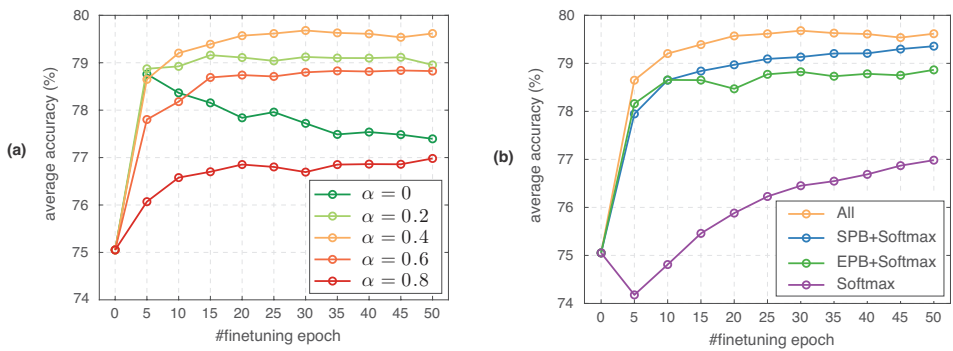}
		\caption{(a) Variation of average accuracy of 75 target subjects during finetuning (\emph{All} strategy) with different values of $\alpha$ . (b) Variation of average accuracy of 75 target subjects during finetuning with different finetuning strategies ($\alpha$ was fixed to $0.4$).} 
		\label{fig:acc_variation}
	\end{figure}

	It should be emphasized again that, different from the regular-setting experiment in Secion \ref{ss:regular_experiment}, only 75 subjects with two recordings were used for the personalization experiment and three subjects with one recording were excluded.
	The effect of KL-divergence regularization in avoiding overfitting for model personalization is exhibited in Figure \ref{fig:acc_variation}(a) when $\alpha$ takes different values in \{0, 0.2, 0.4, 0.6, 0.8\}. Without KL-divergence regularization (i.e. $\alpha = 0$), the average accuracy of the personalized models on 75 target subjects starts declining after 5 finetuning epochs when the models most likely start overfitting the personalization data. The overfitting appears to get worse and worse with ongoing finetuning process as the average accuracy keeps decreasing. When being regularized with KL-divergence (i.e. $\alpha > 0$), the pattern of the average accuracy curve is gradually reversed when $\alpha$ increases, exhibiting a negligible downward tendency when $\alpha=0.2$, plateauing after 25 finetuning epochs with $\alpha=0.4$, and trending upward with larger values for $\alpha$. 
	
	\begin{figure} [!t]
		\centering
		\includegraphics[width=1\linewidth]{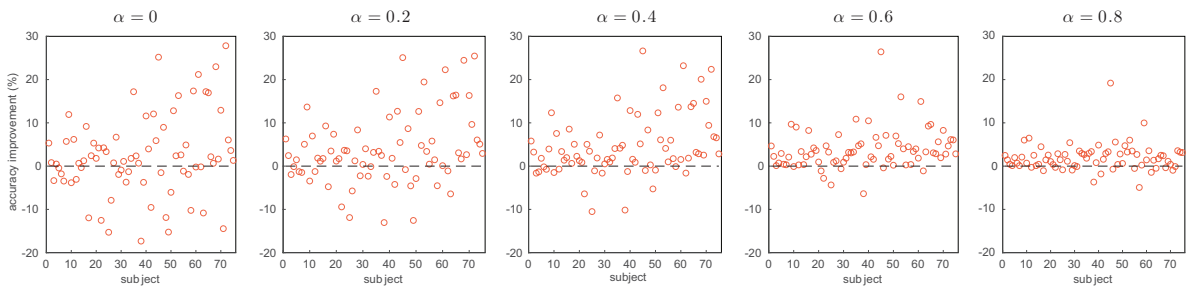}
		\caption{Individual accuracy improvements of 75 target subjects after 50 finetuning epochs when $\alpha$ takes different values in \{0, 0.2, 0.4, 0.6, 0.8\} (\emph{All} finetuning strategy was employed).} 
		\label{fig:gain_scatter}
	\end{figure}

	The results in Figure \ref{fig:acc_variation}(a) also indicate that $\alpha$ plays the role of a trade-off parameter between the pretrained SI model and the purely personal one. When $\alpha$ is set small, we allow the personalized model to aggressively fit to the personalization data at the risk of severe overfitting. In contrast, when $\alpha$ is large, the personalized model is conservatively tied to the SI model and has less freedom to adapt to the personalization data, and so, effectively avoids overfitting at the cost of jeopardizing the personalization. This argument is strengthened with the results in Figure \ref{fig:gain_scatter}. In this figure, the individual accuracy improvements of 75 target subjects varies widely around the baseline zero line when $\alpha=0$ and becomes more and more concentrated towards the zero baseline with increasing value of $\alpha$. Apparently, a value around $0.4$ is a reasonable choice for $\alpha$.
	
\begin{figure} [!t]
		\centering
		\includegraphics[width=0.9\linewidth]{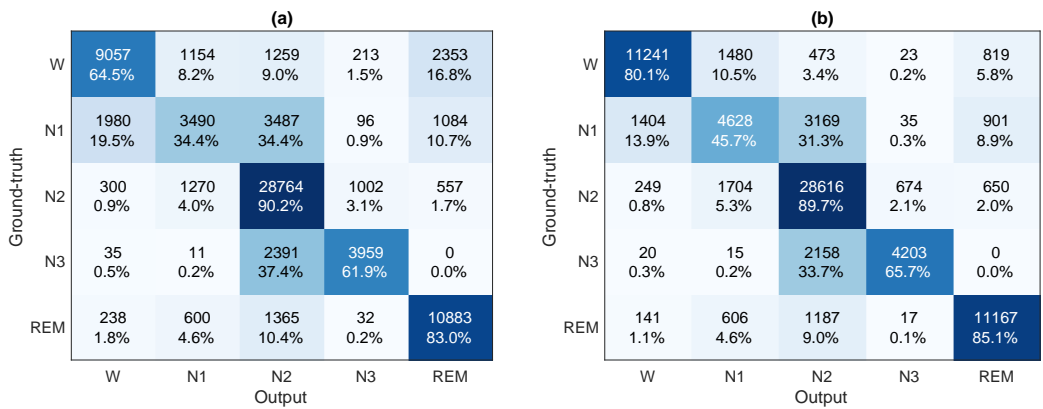}
		\caption{Confusion matrices obtained by SeqSleepNet before and after personalization. (a) Before personalization, (b) After personalization.} 
		\label{fig:personalization_confusion_matrices}
	\end{figure}

	Table \ref{tab:personalization_performance} further provides a comparison of average performance obtained by personalization with different values of $\alpha$ with that before personalization. After personalization, the best performance is obtained with $\alpha=0.4$, reaching an accuracy of $79.6\%$ and improving over that of personalization without KL-divergence regularizaion and that of no-personalization by $2.2$ and $4.5$ percentage points absolute, respectively. Significant improvement on accuracy can also be seen from the confusion matrices in Figure \ref{fig:personalization_confusion_matrices} for most of the sleep stages. Furthermore, this accuracy level is on par with that of the model trained on the entire (known) cohort in Table \ref{tab:scratch_performance} even though only one-night data of the subjects was used and the cohort was unknown.
	
	\setlength\tabcolsep{2.25pt}
	\begin{table}[!t]
		\caption{Average sleep staging performance before and after personalization. Personalization without KL-divergence regularization corresponds to $\alpha=0$ and personalization with KL-divergence regularization corresponds to $\alpha>0$. \emph{All} finetuning was employed and personalization was run for 50 finetuning epochs.}
		\footnotesize
		\begin{center}
			\begin{tabular}{|>{\arraybackslash}m{1in}|>{\centering\arraybackslash}m{0.5in}|>{\centering\arraybackslash}m{0.75in}|>{\centering\arraybackslash}m{0.9in}|>{\centering\arraybackslash}m{0.75in}|>{\centering\arraybackslash}m{0.75in}|>{\centering\arraybackslash}m{0.75in}|>{\centering\arraybackslash}m{0in} @{}m{0pt}@{}}
				\cline{3-7}
				\multicolumn{2}{c|}{} & \multicolumn{5}{c|}{Overall metrics} & \parbox{0pt}{\rule{0pt}{0.25ex+\baselineskip}} \\ [0ex]  	
				
				\cline{3-7}
				\multicolumn{2}{c|}{} & Acc. & $\kappa$ & MF1 & Sens. & Spec. &\parbox{0pt}{\rule{0pt}{0.25ex+\baselineskip}} \\ [0ex]  	
				\cline{1-7}
				\multicolumn{1}{|c}{\makecell{Before \\ personalization}}& & $75.1 \pm 11.2$ & $0.648 \pm 0.140$ & $67.2 \pm 11.4$ & $69.7 \pm 11.4$ & $93.1 \pm 2.8$ &  \parbox{0pt}{\rule{0pt}{0.25ex+\baselineskip}} \\ [0ex]  	
				\cline{1-7}
				\multirow{5}{*}{\makecell{After \\ personalization}} & $\alpha=0~~$ & $77.4 \pm 10.0$ & $0.677 \pm 0.131$ & $71.4 \pm 9.7$ & $69.6 \pm 10.8$ & $93.6 \pm 2.6$ &  \parbox{0pt}{\rule{0pt}{0.25ex+\baselineskip}} \\ [0ex]  	
				& $\alpha=0.2$ & $79.0 \pm 8.4$ & $0.697 \pm 0.114$ & $72.5 \pm 8.9$ & $71.2 \pm 10.2$ & $94.0 \pm 2.3$ &   \parbox{0pt}{\rule{0pt}{0.25ex+\baselineskip}} \\ [0ex]  	
				& $\alpha=0.4$ & $\bm{79.6 \pm 8.4}$ & $\bm{0.706 \pm 0.113}$ & $\bm{73.0 \pm 8.8}$ & $\bm{71.8 \pm 10.1}$ & $\bm{94.2 \pm 2.2}$ &  \parbox{0pt}{\rule{0pt}{0.25ex+\baselineskip}} \\ [0ex]  	
				& $\alpha=0.6$ & $78.8 \pm 10.0$ & $0.697 \pm 0.128$ & $72.0 \pm 10.0$ & $71.6 \pm 10.9$ & $94.0 \pm 2.5$ &  \parbox{0pt}{\rule{0pt}{0.25ex+\baselineskip}} \\ [0ex]  	
				& $\alpha=0.8$ & $77.0 \pm 10.9$ & $0.672 \pm 0.138$ & $69.2 \pm 12.0$ & $70.2 \pm 11.8$ & $93.5 \pm 2.7$ &   \parbox{0pt}{\rule{0pt}{0.25ex+\baselineskip}} \\ [0ex]  	
				\cline{1-7}
			\end{tabular}
		\end{center}
		\label{tab:personalization_performance}
	\end{table}
	
	\subsection{Influence of finetuning strategies.}
	
	It was shown in \cite{Phan2019e,Phan2019f} that, in sleep transfer learning, it is important to finetune feature-learning parts of a pretrained network to overcome the channel mismatch between a source domain and a target domain. This rule of thumb also applies to personalization as shown in Figure \ref{fig:acc_variation}(b). Although finetuning the Softmax component alone brings up the performance, the improvement is significantly lower than the ones obtained by other finetuning strategies in which the feature-learning components of the pretrained SeqSleepNet (i.e. EPB or SBP or both) and the Softmax component are collectively adapted. For instance, the \emph{All} finetuning strategy produces an accuracy improvement of $4.6$ percentage points which is more than twice as much as the $1.9$ percentage points obtained with the \emph{Softmax} finetuning strategy after 50 finetuning epochs.

	\subsection{To personalize or not personalize?}
	\label{ssec:personalizeornot}
	In sleep transfer learning in general, when there is mismatch between the source domain (the MASS databased used for pretraining in our case) and the target domain (the personalization data in our case), it is vital to perform some form of finetuning. In case of personalization, besides possible discrepancies between the source-domain data and the personalization data \cite{Phan2019e}, this data mismatch is further topped up with the target subject's peculiarities. On the contrary, when there is no data mismatch, finetuning could be averted as no significant improvement is expected while one increases the risk of overfitting.
	If there is a way to determine whether data distributions mismatch, one can decide to personalize the sleep staging model or not. Fortunately, we have access to the ground truth of a target subject's one-night data which can be utilized to assess the performance of the pretrained SI model. If the pretrained SI model performs well on this one-night data, the personalization data distribution is very likely matched to the source-domain data distribution. Reversely, poor performance of the pretrained SI model on this personalization data is an indicator of data mismatch.
	
\begin{figure} [!t]
		\centering
		\includegraphics[width=0.8\linewidth]{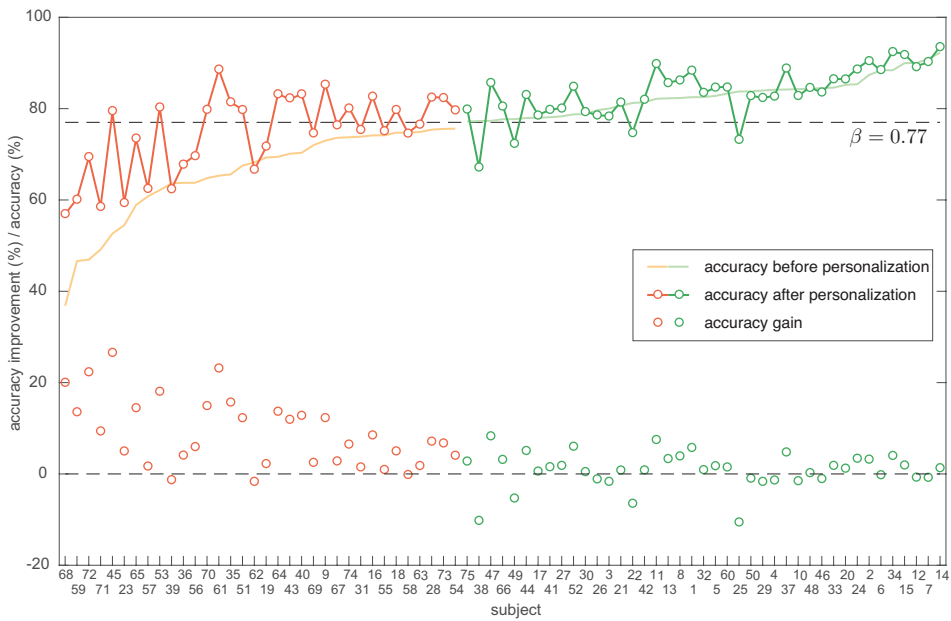}
		\caption{Individual accuracies before personalization, individual accuracies after personalization, and individual accuracy improvements of 75 target subjects (sorted by increasing accuracy before personalization). Group A (i.e. subjects with accuracy before personalization below $\beta$) is marked in orange and Group B (i.e. subjects with accuracy before personalization equal or greater than $\beta$) are marked in green. \emph{All} finetuning was employed, $\alpha$ was fixed to $0.4$, and personalization was run for 50 finetuning epochs.} 
		\label{fig:gain_scatter_sorted}
	\end{figure}

	In light of this observation, we applied a threshold $\beta$ to the individual accuracy obtained on the first-night data to group 75 target subjects into two groups: Group A consisting of subjects with the accuracy before personalization below $\beta$ and Group B consisting of subjects with accuracy before personalization equal or above $\beta$. Figure \ref{fig:gain_scatter_sorted} shows the individual accuracies before personalization, the individual accuracies after personalization, and the individual accuracy improvements of the subjects in both groups with $\beta=0.77$. As can be seen, most significant accuracy improvements correspond to the subjects in Group A while those improvements of the subjects in Group B are much more subtle. On average, personalization for Group A's subjects results in an improvement of $9.0$ percentage points, ten times larger than that for Group B's subjects which is $0.9$ percentage points.
	
	\section{Discussion}
	
	
	
	The personalization results in Figure \ref{fig:gain_scatter_sorted} reveal uneven distribution of accuracy improvement across subjects. Those subjects on which the pretrained SI model performs poorly (i.e. severe data mismatch) benefit the most from personalization. However, only very modest improvements were seen for those subjects on which the pretrained SI model performs well, despite the fact that there is a similar channel mismatch: the C4-A1 EEG channel was used for pretraining the SI model and the Fpz-Cz EEG channel was used for personalization data. We speculate that personalization will be crucial for all target subjects when a completely different channel layout is used, for example in-ear EEG \cite{Mikkelsen2019b} or around-the-ear EEG \cite{Mikkelsen2019,Sterr2018}. 
	
	Setting a right value for the coefficient $\alpha$ was shown to play an important role in personalization's success. Although we have studied  a common $\alpha$ for all target subjects and fixed its value during the personalization process, it makes more sense for $\alpha$ to be adaptive. For example, for subjects with significant peculiarities (e.g. those subjects in Group A in Section \ref{ssec:personalizeornot}), one should start with a large $\alpha$ to impose strong personalization inititally and attenuate it along the ongoing personalization process to gradually reduce this risk. The amount of personalization data should also be taken into account when setting a value for the KL-divergence regularization coefficient $\alpha$. As a matter of fact, using single-night data for personalization is convenient. However, when more data is available, improvement on personalization performance can be expected. In intuition, $\alpha$ should be proportional to the amount of personalization data, i.e. we should use a small $\alpha$ for small personalization data (we trust the SI model more) and a large $\alpha$ for large personalization data (we trust the personalization data more).
	
	\section{Conclusions}
	
	We introduced the problem of sleep-staging personalization with single-night data and discussed its benefits and challenges in the context of personal sleep monitoring. We then attempted to tackle this problem using a transfer learning approach. The subject independent (SI) model (i.e. the pretrained SeqSleepNet) was used as the starting point and finetuned on the single-night data of a target subject to accomplish personalization. KL-divergence between the personalized model's output and the SI model's output is proposed to regularize the network's loss function during personalization. The KL-divergence regularization anchors the personalized model, effectively preventing it from overfitting to the personalization data. Experimenting with 75 subjects of the Sleep-EDF Expanded database, we demonstrated that sleep personalization with a single-night data is possible. We showed that personalization implemented with KL-divergence regularization is robust against overfitting and achieves more favorable results compared to non-personalization and personalization without KL-divergence regularization.
	
	In this pilot study, we demonstrated that automatic sleep staging with single-night data is possible and the obtained results are encouraging. 
	However, while the number of subjects, at 75, is decently high, the population could still be considered quite homogeneous, which could impact the results shown.
	A larger database with diverging and richer characteristics (e.g. demographics, sleep diseases, and electrode placements etc.) is desirable for furture work. Labeling such a database should follow the new and more robust AASM guildlines \cite{Iber2007}.
	
	\section*{Acknowledgment}
	We gratefully acknowledge the support of NVIDIA Corporation with the donation of the Titan V GPU used for this research.
	
	\section*{References}
	\small
	\bibliographystyle{IEEEbib}
	\bibliography{bibliography}
	
\end{document}